\documentclass[10pt,twocolumn,letterpaper]{article}

\usepackage{cvpr}              % To produce the CAMERA-READY version
\usepackage{colortbl}
\usepackage{graphicx}
\usepackage{subcaption}
\usepackage{xspace}
\usepackage{multirow}
\usepackage{makecell}
\usepackage{algorithm}
\usepackage{booktabs}

\usepackage{subcaption}
\usepackage{url}
\usepackage{ragged2e}

\usepackage{caption}
\usepackage{float}
\usepackage{adjustbox}
\newcommand{\bg}{\mathbf{g}}

\newcommand{\bI}{\mathbf{I}}

\newcommand{\bn}{\mathbf{n}}
\newcommand{\bo}{\mathbf{o}}
\newcommand{\bp}{\mathbf{p}}

\newcommand{\br}{\mathbf{r}}\newcommand{\bR}{\mathbf{R}}

\newcommand{\bt}{\mathbf{t}}

\newcommand{\nR}{\mathbb{R}}

\newcommand{\cD}{\mathcal{D}}

\newcommand{\cF}{\mathcal{F}}

\newcommand{\cH}{\mathcal{H}}

\newcommand{\cJ}{\mathcal{J}}

\newcommand{\cL}{\mathcal{L}}

\newcommand{\cP}{\mathcal{P}}
\newcommand{\cQ}{\mathcal{Q}}

\newcommand{\cT}{\mathcal{T}}

\newcommand{\figref}[1]{Figure~\ref{#1}}

\newcommand{\eqnref}[1]{Eq.~\ref{#1}}
\newcommand{\tabref}[1]{Table~\ref{#1}}

\makeatletter
\DeclareRobustCommand\onedot{\futurelet\@let@token\@onedot}
\def\@onedot{\ifx\@let@token.\else.\null\fi\xspace}
\def\eg{e.g\onedot} 
\def\ie{i.e\onedot} 
 
\def\etc{etc\onedot}

\def\etal{et~al\onedot}

\makeatother

\definecolor{yellow}{rgb}{1, 1, 0.7}
\definecolor{orange}{rgb}{1, 0.85, 0.7}
\definecolor{tablered}{rgb}{1, 0.7, 0.7}
\definecolor{red}{rgb}{1, 0, 0}
\definecolor{wincolor}{rgb}{0.85, 0.0, 0.0}
\definecolor{darkyellow}{rgb}{0.8, 0.8, 0.5}
\definecolor{darkred}{rgb}{0.7, 0.3, 0.3}
\definecolor{darkgreen}{rgb}{0.3, 0.7, 0.3}
\definecolor{green}{rgb}{0, 1.0, 0}
\definecolor{pink}{rgb}{1, 0.4, 0.7}
\definecolor{rowcolor}{rgb}{0.902,0.902,0.902}

\definecolor{sred}{rgb}{0.8,0.0,0.0}
\definecolor{sgreen}{rgb}{0,0.8,0}

\def\3d{\textcolor{sred}{\textbf{3D}}}
\def\2d{\textcolor{darkgreen}{\textbf{2D}}}

\definecolor{cvprblue}{rgb}{0.21,0.49,0.74}
\usepackage[pagebackref,breaklinks,colorlinks]{hyperref}

\def\paperID{4072} % *** Enter the Paper ID here
\def\confName{CVPR}
\def\confYear{2025}

\title{3D Prior is All You Need: Cross-Task Few-shot 2D Gaze Estimation}

\author{Yihua Cheng$^1$, \quad Hengfei Wang$^1$, \quad  Zhongqun Zhang$^1$\thanks{Corresponding author}, \quad Yang Yue$^1$, \\ Boeun Kim$^{1,3,4}$, \quad Feng Lu$^2$, \quad Hyung Jin Chang$^1$\\
$^1$University of Birmingham, $^2$Beihang University, $^3$Dankook University, $^4$KETI\\
\url{www.yihua.zone/work/gaze322}
}

\begin{document}
\maketitle

\begin{abstract}
3D and 2D gaze estimation share the fundamental objective of capturing eye movements but are traditionally treated as two distinct research domains. In this paper, we introduce a novel cross-task few-shot 2D gaze estimation approach, aiming to adapt a pre-trained 3D gaze estimation network for 2D gaze prediction on unseen devices using only a few training images. This task is highly challenging due to the domain gap between 3D and 2D gaze, unknown screen poses, and limited training data.
To address these challenges, we propose a novel framework that bridges the gap between 3D and 2D gaze. Our framework contains a physics-based differentiable projection module with learnable parameters to model screen poses and project 3D gaze into 2D gaze. The framework is fully differentiable and can integrate into existing 3D gaze networks without modifying their original architecture.
Additionally, we introduce a dynamic pseudo-labelling strategy for flipped images, which is particularly challenging for 2D labels due to unknown screen poses. To overcome this, we reverse the projection process by converting 2D labels to 3D space, where flipping is performed. Notably, this 3D space is not aligned with the camera coordinate system, so we learn a dynamic transformation matrix to compensate for this misalignment.
We evaluate our method on MPIIGaze, EVE, and GazeCapture datasets, collected respectively on laptops, desktop computers, and mobile devices. The superior performance highlights the effectiveness of our approach, and demonstrates its strong potential for real-world applications.

\end{abstract}    
\section{Introduction}
\label{sec:intro}

Gaze estimation tracks eye movements to predict human attention~\cite{Cheng_2024_pami}. It is a highly applied research topic, where various application scenarios, such as intelligent vehicles~\cite{cheng2024ivgaze, s19245540}, VR/AR~\cite{Palazzi_2019_tpami, patney2016towards, mania2021gaze}, and disease diagnosis~\cite{Wang2022app,bhattacharya2022gazeradar} demand distinct and specialized gaze estimation solutions.

Recent gaze estimation methods primarily focus on 3D gaze estimation~\cite{Zhang_2015_CVPR,cheng2022icpr}, wherein 3D direction vectors are derived from facial images.
Such methods exhibit high adaptability, facilitating straightforward application in diverse environments~\cite{Zhang_2018_CHI}. 
However, they present limitations in practical applications, such as human-computer interaction, where precise gaze targets are essential. Existing approaches often require post-processing to calibrate the pose of interacted objects, e.g., a screen, and compute the intersection between gaze and objects~\cite{Zhang_2019_CHI}. This process poses significant challenges, particularly for non-expert users.

\begin{figure}[t]
	\begin{center}
		\includegraphics[width=\linewidth]{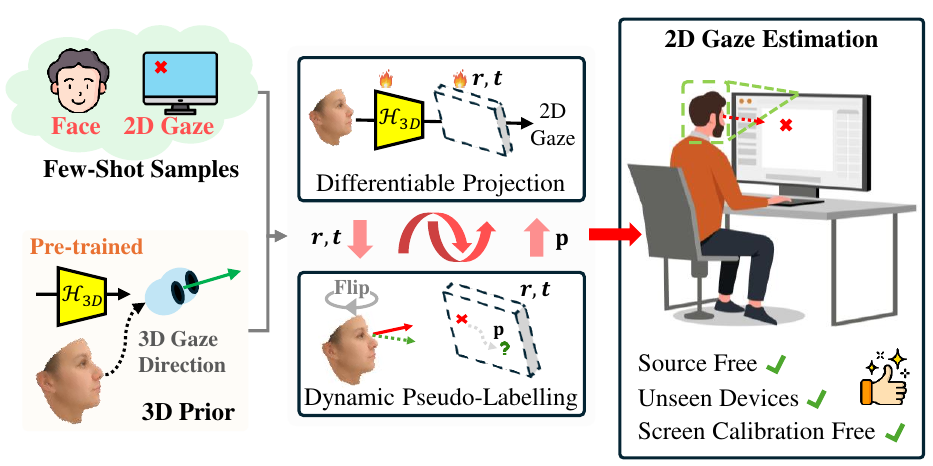}	
	\end{center}
        \vspace{-4mm}
	\caption{We introduce a novel cross-task few-shot 2D gaze estimation approach. Our method leverages a pre-trained 3D gaze estimation network and few-shot 2D gaze samples to achieve 2D gaze estimation on unseen devices. It contains a physics-based differentiable projection module to bridge 3D and 2D gaze, along with a dynamic pseudo-labelling strategy for 2D labels under unknown screen poses. Our approach is both screen-calibration-free and source-free, significantly expanding its application potential.\vspace{-7mm}}
	\label{fig:teaser}
\end{figure}

Conversely, some methods directly estimate 2D gaze within screen coordinate systems. Deep learning-based approaches utilize large training datasets to map facial images to 2D gaze~\cite{Krafka_2016_CVPR,Bao_2020_ICPR}.
However, these models are often entangled with multiple device-specific factors.
Traditional approaches construct 3D eye models using prior anatomical knowledge and fit these models with few-shot calibration images~\cite{hansen2009eye}. Although such methods require specialized equipment for precise eye tracking, they raise the question, \textit{Can we achieve similar patterns within the deep learning paradigm for quick adaptation across various devices? }

In this work, we explore a novel topic, cross-task few-shot 2D gaze estimation. 
We observe that 3D gaze estimation has recently gained significant attention in the research community. It is performed within 3D space, free from entanglement with specific devices.
These insights suggest that 3D gaze estimation models would be a great prior, similar to the 3D anatomical eye model in traditional methods.
Therefore, our approach aims to utilize 3D gaze estimation as prior and adapt it efficiently for 2D gaze estimation. 
However, this setting introduces several significant challenges, such as the domain gap between 3D and 2D gaze tasks, unseen device settings, \ie, \textit{w/o} screen calibration, and insufficient training data. We show a comparison between our task with common methods in \tabref{tab:settingcomparison}. 

To address these challenges, we first propose a novel framework to bridge the gap between 2D and 3D gaze estimation. We decomposes 2D gaze estimation into two components: 3D gaze estimation and gaze projection. 
We first estimate 3D gaze from face images, and then project 3D gaze onto a specific 2D plane to infer the 2D gaze.
Unlike existing methods that require screen calibration to obtain the screen pose~\cite{Cheng_2024_pami,Balim_2023_CVPR}, our framework includes a physics-based differentiable projection module. This module models screen pose using six learnable parameters, \ie, rotation and translation vectors that map screen coordinate system to camera coordinate system. By implementing the projection in a fully differentiable manner, our framework enables seamless integration of the projection module into any existing 3D gaze estimation model without changing its original architecture. 
Furthermore, since the framework is fully differentiable, it supports fine-tuning on 2D annotated data. 

We further propose a dynamic pseudo-labelling strategy for 2D label in our framework. 
Specifically, we perform flipping on face images and aim to assign pseudo-label for the flipped images.
While this process is straightforward for 3D gaze annotations, it becomes more complex for 2D gaze due to dependencies on factors like head position and screen pose, especially the screen pose is unknown. 
To address this, we perform dynamic pseudo-labelling during training. In each iteration, we reverse the projection process using the learnable screen parameters to convert 2D labels into 3D labels. This allows us to perform flipping directly in the 3D gaze space.
A key insight is that flipping needs to occur in the camera coordinate system, while accounting for a shift in coordinate systems during training.
To handle this, we learn a dynamic transformation that maps the shifted system back to the camera coordinate system, ensuring reliable pseudo-label generation.
Additionally, we apply color jittering during training, which does not alter the 2D gaze labels, and minimize uncertainty across jittered images to improve robustness.

\begin{table}[t]
    \arrayrulecolor[rgb]{0,0,0}
    \setlength\tabcolsep{4pt}
    
    \renewcommand\arraystretch{1.0}
    
    \small
    \caption{Comparison between our method with existing methods. We introduce an unexplored task in gaze estimation, which aims to adapt 3D gaze models for 2D gaze estimation with few-shot data. \vspace{-2mm}}
      \centering
        \begin{tabular}{c|ccccc}
        \toprule[0.9pt]
        Category & Train & Test & \makecell{Cross\\Env.} &\makecell{Cross\\Task} & Methods\\
        \hline
        \makecell{3D Gaze Estimation}& \3d &\3d &$\times$&$\times$&\cite{cheng2022icpr,Jindal_2024_CVPR,cheng_2022_aaai}\\
        
        \makecell{2D Gaez Estiamtion}& \2d &\2d &$\times$&$\times$&\cite{Bao_2020_ICPR,Krafka_2016_CVPR}\\
        \multirow{2}{*}{Personalize} & \2d & \2d&$\times$&$\times$&
        \cite{He_2019_ICCV}\\
        &\3d&\3d&$\checkmark$&$\times$&\cite{Lin_2019_iccvw,Park_2019_ICCV}\\
        \makecell{Domain Adaption}& \3d &\3d &$\checkmark$&$\times$&\cite{Cai_2023_CVPR,Bao2_2024_CVPR,Bao_2024_CVPR}\\
        \hline
        Ours& \3d & \2d &$\checkmark$&$\checkmark$&None\\
        \bottomrule[1.0pt]
    \end{tabular}
    \vspace{-4mm}
     \label{tab:settingcomparison}
\end{table}

Overall, our main contribution contains four-folds:

\begin{enumerate}
    
    \item We explore the novel topic of cross-task few-shot 2D gaze estimation. This topic not only extends the application of 3D gaze research to the 2D domain but also provides a promising direction for real-world applications.
    
    \item We propose a framework to bridge 3D and 2D gaze estimation, which includes a physics-based differentiable projection module with six learnable screen parameters to convert 3D gaze to 2D gaze.
    By leveraging this framework, we can quickly adapt a 3D gaze model for 2D gaze estimation using only a small number of images. 

    \item We propose a dynamic pseudo-labeling strategy for 2D labels in our framework. We reverse the projection using learnable screen parameters to convert 2D labels back into 3D labels and perform pseudo-labeling in the 3D gaze space. Furthermore, we learn a dynamic transformation to address the shifted coordinate system problem.

    \item We establish a benchmark for the cross-task few-shot 2D gaze estimation, and evaluate our method on three datasets covering daily scenarios, including laptop, desktop computer and mobile devices. The superior performance demonstrates the advantage of our approach.

\end{enumerate}

\begin{figure*}[t]
	\begin{center}
		\includegraphics[width=2\columnwidth]{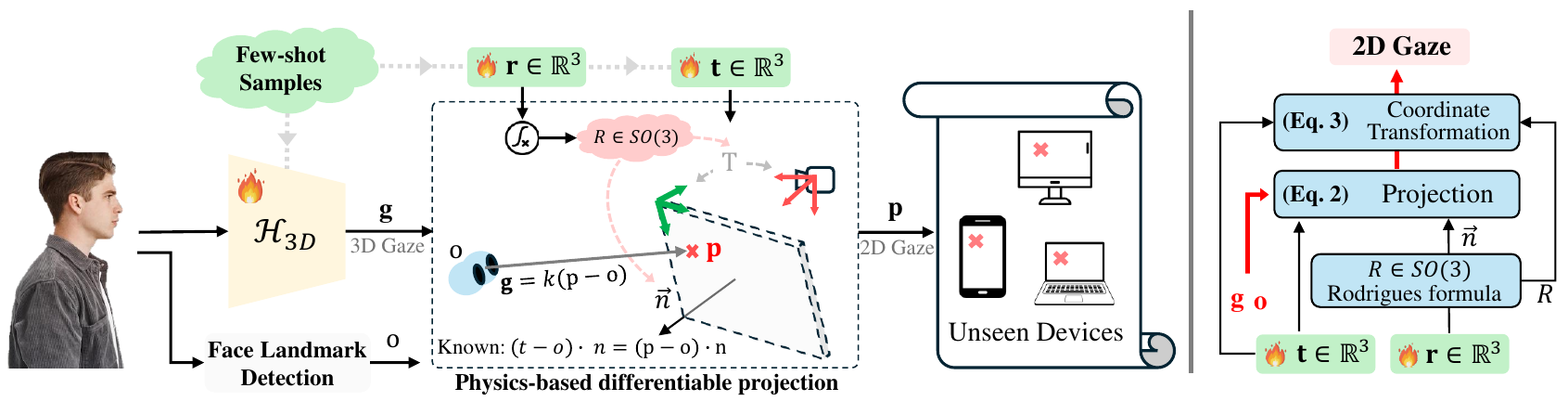}	
	\end{center}
    \vspace{-6mm}
	\caption{We propose a framework for the cross-task few-shot 2D gaze estimation. The framework contains a physics-based differentiable projection module with learnable parameters $\br$ and $\bt$ to model screen, and project 3D gaze into 2D gaze. The framework is fully differentiable and can integrate into existing 3D gaze networks without modifying their original architecture. Leveraging this framework, we can quickly adapt a 3D gaze model for 2D gaze estimation using only a small number of images.\vspace{-6mm}}
	\label{fig:method}
\end{figure*}
\section{Related Works}

\subsection{Gaze Estimation}

Gaze estimation methods are generally classified into 3D and 2D gaze estimation based on output~\cite{Cheng_2024_pami}.  3D gaze estimation defines gaze as a directional vector originating from the face toward gaze targets~\cite{Zhang_2015_CVPR,Cheng_2020_tip}. It typically focuses on enhancing accuracy and generalizability across diverse environments. Related research spans several fields, including supervised learning~\cite{Cheng_2020_AAAI,cheng2022icpr,Cheng_2023_ICCV}, unsupervised domain adaptation~\cite{Cai_2023_CVPR,Bao2_2024_CVPR,Bao_2024_CVPR}, feature disentanglement~\cite{Park_2019_ICCV,cheng_2022_aaai,yin_2024_eccv}, \etc.

On the other hand, 2D gaze estimation is primarily applied in screen-based contexts, where gaze is represented as a pixel coordinate in the screen coordinate system~\cite{Krafka_2016_CVPR,Bao_2020_ICPR,gudi2020efficiency}. Compared to 3D gaze estimation, 2D gaze estimation is more directly applicable to human-computer interaction~\cite{Huynh_2022_ubi, cast_2023_etra, valliappan2020accelerating}. However, it becomes entangled with multiple device-specific factors, such as screen size and camera-screen pose, which complicate generalization across various setups. The adaptation of 2D gaze estimation methods remains a notable research challenge.

Although these gaze estimation methods fundamentally capture eye movement, the distinct differences between 3D and 2D gaze annotations define them as separate research areas. In this work, we propose a framework to bridge the gap between 2D and 3D gaze estimation, enabling the direct application of 3D gaze research to 2D gaze estimation.

\subsection{2D Gaze Estimation via Projection}
It is typical to compute the intersection between 3D gaze and a 2D plane for 2D gaze, a process referred to as gaze projection in this work. The most common application is in VR~\cite{2017_gazevr,2021_gazevr}, where the head-mounted display provides 3D gaze estimation, and developers can easily obtain a plane pose in VR space. They project the gaze onto this plane or determine if it intersects for interaction.

This strategy is also used in deep-learning based gaze estimation methods.
They calibrate the screen pose during the post-processing stage and convert 3D gaze into 2D gaze on the screen~\cite{Cheng_2024_pami,Zhang_2019_CHI}.
Recently, some methods have attempted to inject the projection into the deep learning framework. Balim \etal ~\cite{Balim_2023_CVPR} first require screen calibration to obtain screen parameters and then model the projection process using the calibrated pose.
Cheng \etal~\cite{cheng2024ivgaze} focus on estimating gaze zones on vehicle windshields. They define a basis tri-plane, project 3D gaze onto this plane, and then learn a mapping from the interaction points to the gaze zone.

In our work, we model the full projection process by defining the screen pose with six learnable parameters. The projection module is parameter-efficient. More importantly, our method does not require screen calibration, which can be challenging for non-expert users.

\section{Methodology}

\subsection{Task Definition}
Given a pre-trained 3D gaze estimation network $\cH_{3D}\left(\bI;  \beta\right)$, which takes face images $\bI$ as input and outputs 3D gaze direction $\bg$, \ie, $\cH_{3D}:\bI\rightarrow\bg$,
our objective is to develop a  2D gaze estimation network $\cH_{2D}\left(\bI;  \theta\right)$. Using few-shot training samples $\cD = \{(\bI_i, \bp_i)\}_{i=1}^N$, where $N$ is the number of training samples, this network estimates 2D pixel coordinates $\bp$ from face images, \ie, $\cH_{2D}:\bI \rightarrow\bp$. We consider a restricted setting where: 1) the method is source-free, as the training set of $\cH_{3D}$ is unavailable, and 2) it is screen calibration-free, with the screen pose unspecified. These restrictions make our method convenient for practical applications while upholding data privacy.

\subsection{Physics-Based Differentiable Projection}

Our work aims to learn a new $\cH_{2D}$ using few-shot samples.
The primary challenge lies in transferring knowledge from $\cH_{3D}$ to $\cH_{2D}$. However, the two networks perform different tasks, making some conventional methods such as feature distillation unsuitable.
To address this, our idea is to decompose the 2D gaze estimation into 3D gaze estimation and gaze projection. Specifically, we incorporate $\cH_{3D}$ as part of $\cH_{2D}$, supplemented with an additional module for projecting gaze directions onto a 2D screen.
Unlike existing gaze projection strategies that often rely on post-processing~\cite{Cheng_2024_pami} or require screen calibration~\cite{Balim_2023_CVPR}, we introduce a physics-based differentiable projection module. This module models screen pose as learnable weights, enabling the projection process to occur in a differentiable and adaptable manner.

In detail, we define learnable weights $\br\in\nR^3$ as the rotation vector and $\bt\in\nR^3$ as the translation vector within the projection module,
establishing the transformation from the screen coordinate system to the camera coordinate system.
To transform $\br$ into the rotation matrix $\bR\in\nR^{3\times3}$, we apply the Rodrigues formula, which preserves the orthogonality of $\bR$ so that $\bR \in SO(3)$.
The input of projection module contains gaze direction $\bg\in\nR^3$ and the 3D position of the face center $\bo\in\nR^3$, the latter of which can be computed using existing 3D landmark estimation methods~\cite{Guo_2020_ECCV}. Overall, the module $\cP$ could be denoted as:
\begin{equation}
    \hat{\bp} = \cP(\bg, \bo; \br, \bt), 
\end{equation}
\noindent where $\hat{\bp}\in\nR^2$ represents the estimated screen coordinate. 

We first compute the intersection points between gaze directions and the learnable screen denoted with $(\br, \bt)$. To establish the screen pose, we need a normal vector $\bn$ and a point coordinate on the screen. The normal vector can be derived using $\bR[:,2]$, \ie, the third column of the rotation matrix~\cite{Cheng_2024_pami}, while $\bt$ serves as a reference point on the plane. 
Given that the dot product between the normal vector of a plane and the vector connecting any point on the plane to a fixed point is constant, it is obvious that the intersection point $\bp_{3D}$ is

\begin{equation}
    \bp_{3D} = \bo + \frac{(\bt - \bo)\cdot{\bn}}{\bg\cdot{\bn}} \bg,
\end{equation}

Note that $\bp_{3D}$ represents coordinates in the camera coordinate system. To convert it to the screen coordinate system, we apply
\begin{equation}
    \bp = \bR^{-1}(\bp_{3D} - \bt),
    \label{equ:rot}
\end{equation}

\noindent We slightly abuse the notation $\bp$ in \eqnref{equ:rot}, where the final 2D gaze coordinate corresponds to the first two components of $\bp$. These values can then be further converted into pixel coordinates by utilizing the screen's PPI (pixels per inch), which is easily obtainable as a screen parameter.

Therefore, the network $\cH_{2D}$ can be denoted as 
\begin{equation}
    \cH_{2D}(\bI, \bo; \beta, \br, \bt) = \cP(\cH_{3D}(\bI), \bo),
\end{equation}
\noindent and the objective function is denoted as 
\begin{equation}
    \min_{\beta, \br, \bt} \sum_{i=1}^{N}\left\|\cH_{2D}(\bI_i, \bo_i) - \bp_i\right\|_1  
    \label{equ:loss1}
\end{equation}
\noindent We illustrate the projection module in \figref{fig:method}.

\subsection{Dynamic Pseudo-Labeling for 2D Gaze}
Data augmentation is a typical technique to improve model performance, particularly with limited dataset sizes. In this section, we apply flipping to expand the data space.
In 3D gaze estimation, the flipping involves horizontally flipping face image and adjusting the label by negating the x-coordinate value. We formally define the operation in label as $\cF(\bg)$.
However, generating reliable pseudo labels after flipping is challenging for 2D gaze estimation. 

Our core idea is to dynamically generate pseudo-labels during training by leveraging the differentiable projection module within our framework, which includes learnable screen parameters. 
This enables us to address the challenges of assigning 2D pseudo-labels by reversing the projection process, \ie, converting 2D screen coordinates into 3D gaze directions, where we can then apply flipping in 3D space. The pseudo-labeling function $\cQ(\bp)$ is defined as 
\begin{equation}
    \cQ(\bp) = \cP(\cF(\cP^{-1}(\bp))),
    \label{equ:flip}
\end{equation}
\noindent where $\cP^{-1}$ represents the reverse projection process.
Specifically, we first transform $\bp$ into the camera coordinate system. The gaze direction is then defined as the vector originating from the face center and directed toward the gaze point,

\begin{equation}
    \cP^{-1}(\bp, \bo) = (\bR\bp + \bt) - \bo,
\end{equation}
\noindent which we normalize to ensure the vector has a unit length.

However, we observed that assigning pseudo-labels as \eqnref{equ:flip} led to model collapse, with the pseudo-labels diverging to large values during training.
On the other hand, we found that $\cH_{2D}$ struggled to learn the correct screen parameters, and noted substantial changes in the 3D gaze estimation network itself.
Our intuition suggests that \textit{changing the screen pose should theoretically allow us to find an optimal screen pose, but this could also be approached by rotating the camera instead, \ie, optimize $\cH_{3D}$.}

Based on the observations, we find that \eqnref{equ:flip} is not consistently reliable.
The key insight is that human gaze direction is inherently defined in the camera coordinate system. Flipping image affects the camera coordinate system itself, meaning the gaze label should be adjusted accordingly, \ie, \textit{flipping should be performed in camera coordinate system}. However, since the 3D gaze estimation network undergoes updates during fine-tuning, it is shifted into an unknown coordinate system.
This change in coordinate systems disrupts the alignment of gaze labels, leading to model collapse.

\begin{figure}[t]
	\begin{center}
		\includegraphics[width=\columnwidth]{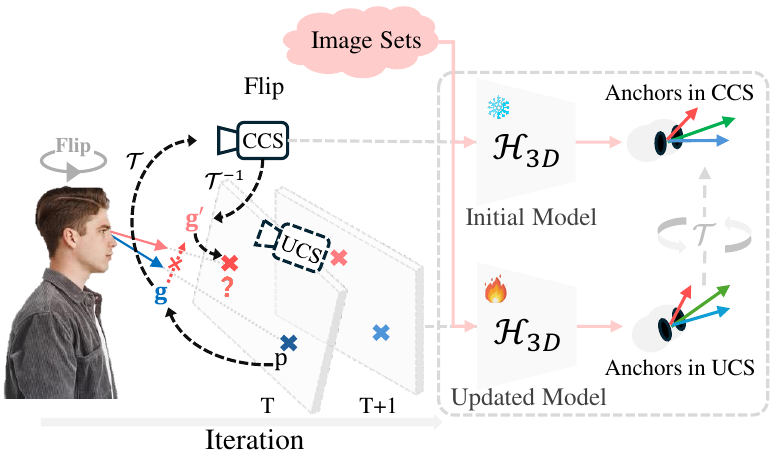}	
	\end{center}
   \vspace{-3mm}
	\caption{
    The dynamic pseudo-labeling strategy for 2D gaze involves reversing the projection process to convert 2D gaze into 3D space, where we compute pseudo-labels.
    To align the camera coordinate system (CCS) with the unknown coordinate system (UCS), we use the same image sets as input to both the initial and the updated 3D model. The initial model, trained on the CCS, while the updated model operates within the UCS. By leveraging the outputs from these models as two anchors, we derive the transformation $\cT$ to align the coordinate systems. Notably, $\cT$ should be invertible. \vspace{-6mm}}
	\label{fig:calibration}
\end{figure}

To solve this problem, we aim to learn a transformation $\cT$ that maps the unknown coordinate system to camera coordinate system.
Our idea is to identify anchors in the two coordinate systems, allowing us to model this problem as an alignment task.
Specifically, we denote the initial pre-trained 3D gaze estimation network as $\cH_{3D}(;\beta_0)$ and the fine-tuned network as $\cH_{3D}(;\beta_k)$.
Notably, $\cH_{3D}(;\beta_0)$ is pre-trained in the camera coordinate system, while $\cH_{3D}(;\beta_k)$ operates in the unknown coordinate system. Therefore, we can acquire the anchor as $\{\cH_{3D}(\bI_i;\beta_0)\}_{i=1}^N$ and $\{\cH_{3D}(\bI_i;\beta_k)\}_{i=1}^N$ using training set $\cD$. The alignment problem can then be formulated as:
\begin{equation}
    \min_\cT \sum_{i=1}^N \|\cT\cH_{3D}(\bI_i;\beta_k) - \cH_{3D}(\bI_i;\beta_0)\|_2
\end{equation}
Notably, $\cT$ should be invertible. Therefore, we model the transformation as a rotation operation, enabling us to solve it using singular value decomposition (SVD). We have 
\begin{equation}
    [U, S, V] = \mathrm{SVD}(\cH_{3D}(\bI_i;\beta_i)*\cH_{3D}(\bI_i;\beta_0)^T),
\end{equation}
and $\cT = VU^T$. Consequently, we can update \eqnref{equ:flip} as follows
\begin{equation}
    \cQ(\bp) = \cP(\cT^{-1}*\cF(\cT*\cP^{-1}(\bp))),
    \label{equ:newflip}
\end{equation}
\noindent where $*$ represents matrix multiplication.
$\cQ(\bp)$ is also dynamic and re-computed during each iteration since the coordinate system continues to change throughout fine-tuning.

The objective function is denoted as 
\begin{equation}
    \min_{\beta, \br, \bt} \sum_{i=1}^{N}\left\|\cH_{2D}(\bI_i', \bo_i) - \cQ(\bp_i)\right\|_1
    \label{equ:loss2}
\end{equation}
Where $\bI_i'$ is the flipping image of $\bI_i$.

\subsection{Minimize Uncertainty across Jittered Images}
We also perform color jitter and minimize uncertainty across jittered images to enhance model robustness. 
Given a face image $\bI$, we apply color jitter $\cJ$ to create a set of augmented images, $\{\cJ_k(\bI)\}_{k=1}^K$, where $k$ represents the number of random color jitters performed. We minimize the variance in the gaze predictions for this set. Specifically, we pass each augmented image through the model, obtaining predictions $\{\cH_{2D}(\cJ_k(\bI))\}_{k=1}^K$.
We calculate the centroid of these predictions and minimize the distance between each prediction and the centroid. Additionally, we also minimize the distance between the predictions and the ground truth.
To stabilize training, we introduce a temporal weight $\tau=\frac{t-1}{t}$ for the variance loss, starting with a smaller weight that increases over epoch $t$. The loss is defined as 
\begin{equation}
\begin{aligned}
    \cL_{unc} = & \frac{1}{NK} \sum_{i=1}^N\sum_{k=1}^K (\|\cH_{2D}(\cJ_k(\bI_i)) - \bp_i\|_1 + \\
    &\tau*\|\cH_{2D}(\cJ_k(\bI_i)) - \frac{1}{K}\sum_{j=1}^K\cH_{2D}(\cJ_j(\bI_i))\|_2)
\end{aligned}
\label{equ:loss3}
\end{equation}
The temporal weight mitigates the risk of model collapse, as we observe that the second term of $\cL_{unc}$ tends to be large at the start of training, and a high initial learning weight can lead to instability. Additionally, we apply L2 regularization  to the second term since it assigns greater weight to outliers.

\subsection{Implementation Details}
Our model is optimized using the loss functions defined in \eqnref{equ:loss1}, \eqnref{equ:loss2} and \eqnref{equ:loss3}, with corresponding weights of 1, 0.4, and 0.25, respectively.
For training, we set $N=10$, meaning the training set contains 10 samples, and $K=4$, meaning we apply four random color jitter augmentations per iteration. The model is implemented in PyTorch and trained on an NVIDIA RTX 3090. We train for 80 epochs, setting the learning rate initially to 0.001, with a 5-epoch warmup phase. After 60 epochs, the learning rate decays to 0.0005. We use GazeTR~\cite{cheng2022icpr} (ResNet18 + 6-layer transformer) pretrained on Gaze360~\cite{Kellnhofer_2019_ICCV} as the basic 3D model. 
Please refer the supplementary material for more details.

\section{Experiment}

\subsection{Setup}
In this paper, we propose a cross-task few-shot 2D gaze estimation task. We first build the evaluation benchmark.

\noindent\textbf{Datasets:} We evaluate methods on three datasets: MPIIGaze~\cite{Zhang_2017_CVPRW}, EVE~\cite{park_2020_eccv}, and GazeCapture~\cite{Krafka_2016_CVPR}. These datasets were collected in different devices, including laptops, desktop computers, and mobile devices. By assessing performance across these datasets, we demonstrate the generalization capability of methods across various devices.

\noindent\textbf{Data Preprocessing:}
Image normalization~\cite{Cheng_2024_pami} is usually used to enhance 3D gaze estimation performance.  In our work, we utilize the normalized images provided by the MPIIGaze and EVE datasets, and implement the method~\cite{Zhang_2018_etra} for normalizing the GazeCapture. Note that, the normalization changes 3D gaze with a rotation matrix.
Although our work does not use the 3D label, the predicted 3D gaze should be transformed back for projection.
Furthermore, the MPIIGaze dataset augments 3D gaze estimation data by flipping images, which is not applicable for 2D gaze estimation. We exclude the flipped images for consistency.
The EVE dataset provides videos along with corresponding gaze trajectories. We  sample one frame for every 20 frames to construct the benchmark.
We sample 20 subjects in GazeCapture dataset, ensuring that each has at least 500 images. We clean the dataset to remove images without face. Notably, four of the 20 subjects used a tablet for data collection, while the rest used phones.
Please refer the supplementary materials for more details.

% 27 in GC, 33 in EVE and 43% in MPIIGaze
\begin{table}[t]
    \arrayrulecolor[rgb]{0,0,0}
    \setlength\tabcolsep{1pt}
    
    \renewcommand\arraystretch{1.0}
    \small
    \caption{Quantitative evaluation. Our method achieves best result among comparison methods. We also report the performance of 2D gaze estimation methods in the second row for reference.\vspace{-2mm}}
      \centering
        \begin{tabular}{l|cccc}
        \toprule[1.0pt]
       Method & \makecell{Training\\ Samples} & \makecell{EVE\\\cite{park_2020_eccv}} &\makecell{MPIIGaze\\\cite{Zhang_2015_CVPR}}  & \makecell{GazeCapture\\\cite{Krafka_2016_CVPR}}\\
        \hline
        iTracker~\cite{Krafka_2016_CVPR}&\multirow{5}{*}{\makecell{All \\dataset}}&-&-&26.8\\
        EyeNet~\cite{park_2020_eccv}&&49.7&-&-\\
        Full-Face~\cite{Zhang_2017_CVPRW}& &38.6& 42.0&-\\
        AFF-Net~\cite{Bao_2020_ICPR}& &-&39.0&19.6\\
        EFE~\cite{Balim_2023_CVPR} & & 38.5& 38.9&20.5\\
        
        \hline
        EFE~\cite{Balim_2023_CVPR}&10& 64.9 \textcolor{red}{$\blacktriangledown 33\%$ } & 100.2 \textcolor{red}{$\blacktriangledown 43\%$ }& 48.5 \textcolor{red}{$\blacktriangledown 26\%$ } \\
        IVGaze~\cite{cheng2024ivgaze}& 10&177.7 \textcolor{red}{$\blacktriangledown 75\%$ } &132.2 \textcolor{red}{$\blacktriangledown 57\%$ }&68.1 \textcolor{red}{$\blacktriangledown 47\%$ }\\
        \rowcolor{rowcolor}Ours&10&43.4 & 56.7& 35.7\\
        \bottomrule[1.0pt]
    \end{tabular}
     \label{tab:exp1}
     \vspace{-4mm}
\end{table}

\noindent\textbf{Evaluation Metric:}
We perform person-specific evaluation and report the average performance across subjects for comparison. 
Performance is measured as the Euclidean distance (in mm) between predictions and ground truth, where lower values indicate better accuracy.

%-------------------------------------------------------------------------

\subsection{Quantitative Comparison}
We first compare our method with existing approaches EFE~\cite{Balim_2023_CVPR} and IVGaze~\cite{cheng2024ivgaze}. EFE is an end-to-end gaze estimation method that includes a projection module to convert 3D gaze predictions into 2D gaze. IVGaze utilizes a basis tri-plane for projection, followed by a lightweight transformer to refine the projection points.
For a fair comparison, we re-implement both methods using the same 3D gaze estimation network and pre-trained weights as our method. Our goal is to evaluate the performance differences resulting from different projection strategies. Notably, EFE requires screen calibration for the projection; to ensure fairness, we set these screen parameters as learnable and initialize them with the same values used in our method.
The results of these comparisons are presented in \tabref{tab:exp1}.

IVGaze includes a transformer to refine projection points. While this transformer performs well when trained on the full dataset, it struggles with limited data, leading to underfitting when trained on just 10 samples. This results in poor performance on the EVE and MPIIGaze datasets, highlighting the advantage of our approach.
In contrast, our method avoids the use of complex architectures that can suffer from underfitting in few-shot learning tasks. Instead, we directly model the projection process, leading to superior performance.
On the other hand, EFE demonstrates reasonable performance, but our method achieves over $25\%$ improvement across all three datasets. This significant boost is attributed to our more comprehensive modelling of the projection process, which reduces fitting complexity and naturally enhances overall performance.

We also report the performance of 2D gaze estimation methods trained on the entire dataset for reference. Note that they are not directly comparable to our method since both the training and test sets differ. These results are summarized in the second row of \figref{tab:exp1}. Our method achieves similar performance using only 10 images.                                                               
\begin{table}[t]
    \arrayrulecolor[rgb]{0,0,0}
    \setlength\tabcolsep{4pt}
    
    \renewcommand\arraystretch{1.0}
    \small
    \caption{Comparison with different 3D to 2D adaption strategy. We direct project 3D gaze to 2D gaze using the known screen pose without fine-tuning, which shows the advantage of our learning framework. We directly learn 2D gaze from 3D gaze with MLP, which highlights the challenges in the adaption from 3D model to 2D gaze estimation. We also show the performance when the learnable parameters is set as known pose in our method.\vspace{-2mm} }
      \centering
        \begin{tabular}{l|cccc}
        \toprule[1.0pt]
         Strategy& EVE &MPIIGaze &GazeCapture\\
        \hline
        Direct Projection &80.5& 101.9 & N/A\\
        % Direct Projection ($\leq$ 1500) &85.3 &78.1&N/A\\
        Direct Learning  &180.6&133.9&74.23\\
        Direct Learning (with $\bo$ ) &116.6&108.2&149.7\\
        \hline
        Learning with Known Pose&39.4&56.6&N/A\\
        \rowcolor{rowcolor}Ours& 43.4& 56.7&35.7\\
        \bottomrule[1.0pt]
    \end{tabular}
    \vspace{-2mm}
     \label{tab:diffproj}
\end{table}

\subsection{Comparison with Different Adaption Strategy}
In this section, we evaluate the accuracy of different adaption strategies for obtaining 2D gaze from 3D predictions.

\begin{table}[t]
    \arrayrulecolor[rgb]{0,0,0}
    \setlength\tabcolsep{4pt}
    
    \renewcommand\arraystretch{1}
    
    \small
    \caption{We perform an ablation study to evaluate the impact of the dynamic pseudo-labeling strategy (PS-Label) and the loss to minimize uncertainty across jittered images ($\mathcal{L}_{unc}$). Both the two modules contribute to performance improvements.\vspace{-2mm}
    }
      \centering
        \begin{tabular}{ccc|cccc}
        \toprule[1.0pt]
        Proj. &PS-Label & $\cL_{unc}$ & EVE &MPIIGaze &GazeCapture\\
        \hline
      % \checkmark & & &  & 17.76&13.22 &29.58\\
       \checkmark&&&46.6 & 60.3 &36.8 \\
        \checkmark &\checkmark& &45.3 &57.9 & 35.7 &  \\
         \checkmark&\checkmark&\checkmark&43.4 & 56.7& 35.7\\
        \bottomrule[1.0pt]
    \end{tabular}
     \label{tab:ab}
     \vspace{-2mm}
\end{table}

\noindent \textbf{Direct Projection:}
We directly project the 3D gaze predictions from our pre-trained 3D gaze estimation network onto the screen using the known screen pose, providing a baseline performance measure for the network. This is not performed on GazeCapture, as it lacks reliable screen pose.

\noindent \textbf{Direct Learning:}
We retain the architecture of the 3D gaze network and directly fine-tune it using the 2D annotations. Additionally, we concatenate the gaze origin $\bo$ with the predicted gaze and use a MLP to map them to 2D gaze predictions. We then fine-tune this extended network and report the performance in Direct Learning (with $\bo$).

\noindent \textbf{Learning with Known Pose:}
 Our method assumes the screen pose is unavailable. In this strategy, we change the learnable parameters as the ground truth screen pose.

The result is shown in \tabref{tab:diffproj}. 
The Direct Projection method struggles to perform effectively on the EVE and MPIIGaze datasets without fine-tuning. However, integrating it into our framework yields over $40\%$ improvement, demonstrating the critical role of our learning framework.
The Direct Learning strategy, on the other hand, fails to achieve reasonable performance due to the substantial domain gap between 3D and 2D gaze estimation. 
We compare its performance with Direct Projection. The learning strategy does not show any performance gains, which highlights the challenge of adapting 3D gaze models to 2D tasks.
Even when the gaze origin is included as an additional feature, the limited training data makes it challenging for the model to learn the complex mapping. In contrast, our framework leverages physics-based differentiable projection, enabling it to achieve superior performance.
The Learning with Known Pose method outperforms our method due to access to the known screen pose, highlighting the importance of accurate screen pose information for 2D gaze estimation.

\begin{figure}[t]
	\begin{center}
		\includegraphics[width=\columnwidth]{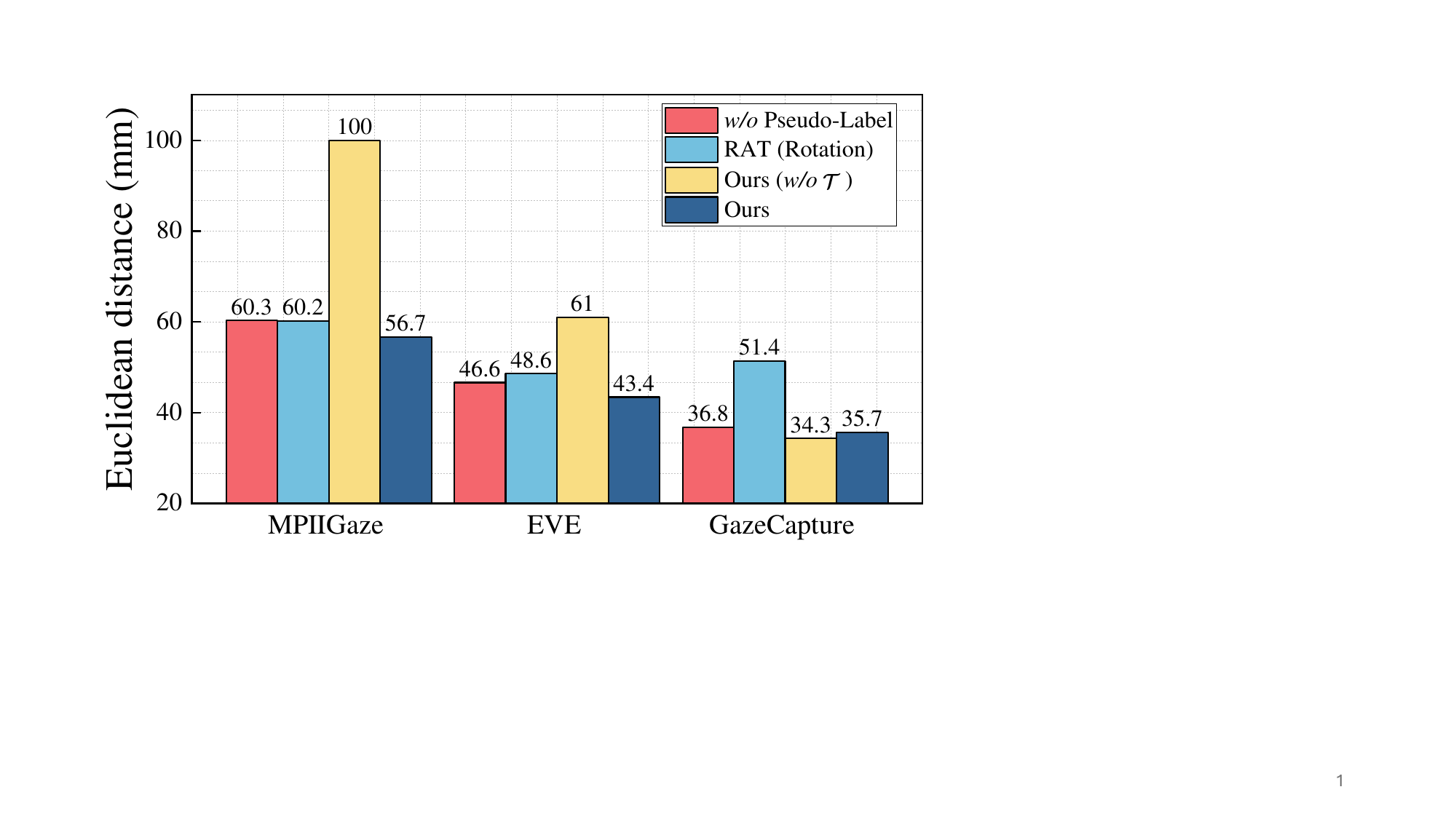}	
	\end{center}
    \vspace{-4mm}
	\caption{We compare the performance across different pseudo-labelling strategies. The red bar represents the projection without pseudo-labelling, serving as a baseline for comparison. We evaluated our method without the transformation $\cT$. The unreliable pseudo-labels lead to significant performance drop on the MPIIGaze and EVE. Interestingly, omitting $\cT$ led to improved results on the GazeCapture dataset. We found that this was because the initial screen pose happened to be same as the actual screen pose. \vspace{-4mm}}
	\label{fig:plcom}
\end{figure}

\subsection{Ablation Study}
We perform an ablation study to demonstrate the contribution of each module in our work. We first evaluate the performance when only the projection module is added to the pre-trained 3D gaze estimation network and fine-tuned. The results are shown as \textit{Proj.} in \tabref{tab:ab}. Compared to the results in \tabref{tab:diffproj}, the projection module provides a significant performance improvement as it explicitly modelling the projection process, which effectively bridges the gap between 3D and 2D gaze estimation.
Next, we introduce our dynamic pseudo-labeling strategy and minimize the uncertainty across jittered images. Both mechanisms bring performance improvements across all datasets.

The dynamic pseudo-labeling strategy is a key contribution of our work. To better understand its impact, we conduct a detailed comparison, as shown in \figref{fig:plcom}. We perform an ablation on the learning transformation $\mathcal{T}$ in our strategy. The results show a significant performance drop on the MPIIGaze and EVE datasets without $\mathcal{T}$, as unreliable pseudo-labels can cause model collapse during learning, especially with small training dataset sizes.
Interestingly, we observe improved performance on the GazeCapture dataset without using $\mathcal{T}$. The authors of GazeCapture create a unified prediction space for 2D gaze, centered at the phone camera position. Our model initializes the screen pose as $\bt = (0, 0, 0)$, making the initial pose closely approximate the real one. However, it is important to note that such cases are uncommon in real-world scenarios.
Our method first converts 2D gaze to 3D space and learns $\mathcal{T}$ to align this space with the camera coordinate system. When the screen pose aligns exactly with ground truth, the 3D space already corresponds to the camera coordinate system. To establish the alignment, we use the predictions from the 3D gaze network as anchors for the camera coordinate system, which may introduce some bias. Nonetheless, our method demonstrates performance improvements compared to methods without pseudo-labeling.

We also implement existing method RAT~\cite{Bao_2022_CVPR}, which assigns pseudo-label for rotated images.
We convert 2D gaze into 3D gaze using learnable screen parameters, and perform RAT to augment training. RAT cannot bring performance improvement compared with the baseline.

\begin{figure}
    \begin{minipage}{0.5\linewidth}
        \centering
        \includegraphics[width=\linewidth]{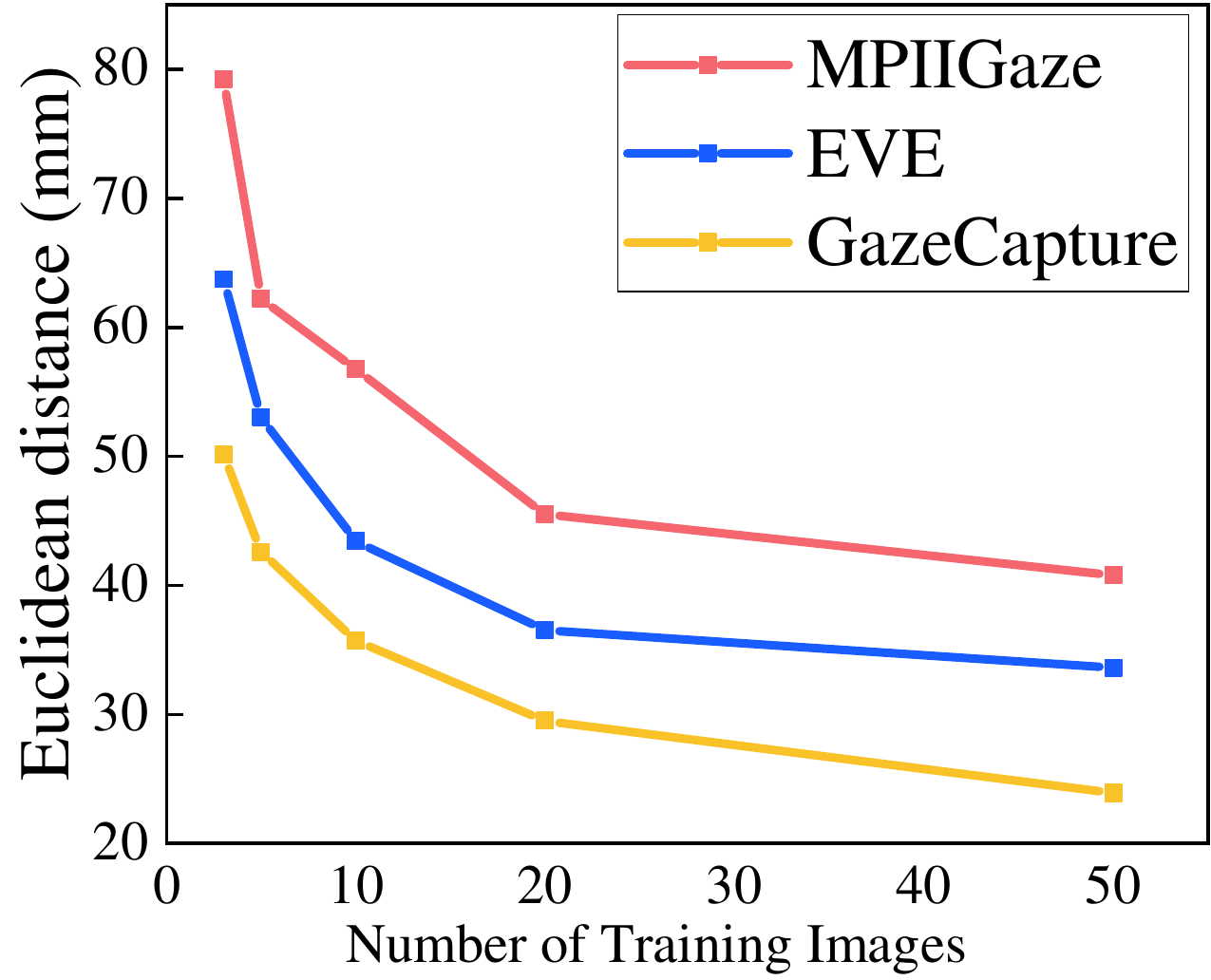}
        \caption{Performance with different number of training images.}
        \label{fig:num}
    \end{minipage}
    \quad
    \begin{minipage}{0.43\linewidth}
    \centering
    \resizebox{\linewidth}{!}{
            \begin{tabular}{|c|c|}
                \hline
                \makecell{\#Training\\Images} & \makecell{Speed\\(sec/epoch)} \\
                \hline
                3 & 0.89 \\
                5 & 0.90 \\
                10 & 0.91 \\
                20 & 0.96 \\
                50 & 1.16 \\
                \hline
            \end{tabular}
    }
    \captionof{table}{The model training time with different number of training samples.}
    \label{tab:num}
\end{minipage}
\vspace{-5mm}
\end{figure}

\subsection{Different Numbers of Training Images }
In this section, we evaluate the effect of the number of training images on model performance. We experiment with different numbers of training images set to 3, 5, 10, 20, and 50, respectively. The performance is assessed across all three datasets, with results depicted in \figref{fig:num}. As shown, increasing the number of training images consistently improves the model performance.

Additionally, we measure the model training time when using varying numbers of training images, as summarized in \tabref{tab:num}. On average, each epoch takes approximately 0.9 seconds to process. Since our method does not require a large dataset, all images can be efficiently processed within a single epoch. With a total of 80 epochs, the complete training time is approximately 1.2 minute. Notably, this timing was tested in a Python environment and could be further optimized to achieve even faster performance with specific optimizations. This demonstrates significant real-time application potential for our method.

\subsection{Repeatability Experiment}
\begin{figure}[t]
	\begin{center}
		\includegraphics[width=\columnwidth]{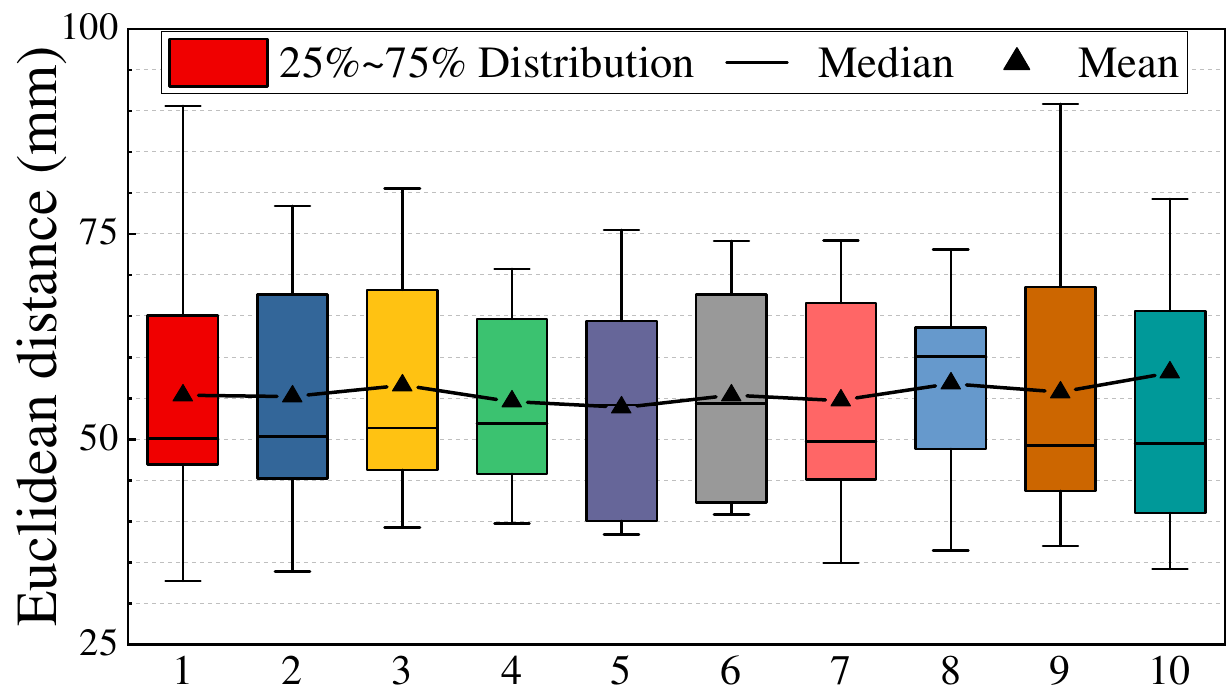}	
	\end{center}
    \vspace{-2mm}
	\caption{We train our method 10 times using different image samples in MPIIGaze for robustness evaluation.  The horizontal axis corresponds to each of the 10 trials, while each bar shows the accuracy distribution across 15 subjects.  The box depicts the interquartile range ($25\%$ to $75\%$), while the error bars covers the entire accuracy distribution. The average accuracy across  10 trials is $55.6$, demonstrating the stability and robustness of our method. \vspace{-6mm} }
	\label{fig:repeat}
\end{figure}

In this section, we conduct a robustness evaluation by training our method 10 times using different training samples in MPIIGaze to assess the impact of sample variability on model performance. We evaluate the performance on all 15 subjects for each trial and report the performance distribution. The results are visualized in a boxplot in \figref{fig:repeat}.
The horizontal axis represents each of the 10 trials and each trial contains performance of 15 subjects. The box depicts the interquartile range ($25\%$ to $75\%$), while the error bars cover the entire performance distribution. The triangle symbol indicates the average performance, and the black line represents the median performance.
The average performance across all 10 trials is $55.6$, which is slightly better than our previously reported value of $56.7$. These results demonstrate the stability and robustness of our method despite variations in the training samples.

\subsection{The Trajectories of Pseudo-Label}
\begin{figure}[t]
	\begin{center}
		\includegraphics[width=\columnwidth]{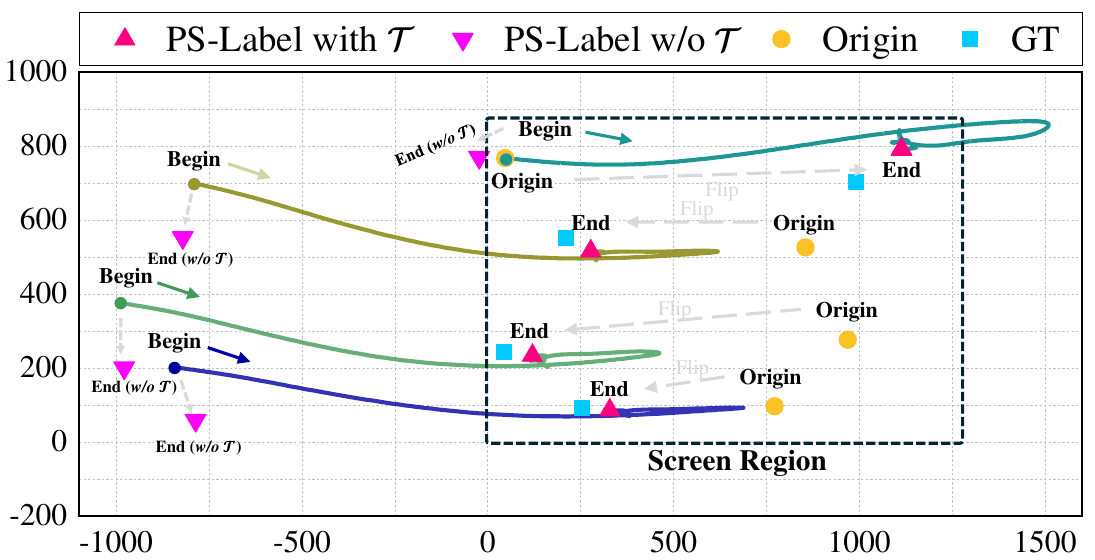}	
	\end{center}
    \vspace{-4mm}
	\caption{ We visualize four trajectories of the pseudo-labels in our dynamic pseudo-labeling strategy. The ground truth for flipped images is computed using known screen pose. It is evident that our method progressively aligns the pseudo-labels closer to the ground truth. Additionally, we plot the pseudo-labels without applying $\cT$ in our strategy, which shows a failure to produce reliable pseudo-labels, resulting in significant deviations from the ground truth. \vspace{-4mm}}
	\label{fig:trajectory}
\end{figure}

Our method contains a dynamic pseudo-labeling strategy to assign pseudo 2D labels for flipped images. To gain deeper insights into this process, we visualize the trajectories of the pseudo-labels over the course of 80 epochs in \figref{fig:trajectory}.
In addition, we compare the effect of our transformation strategy by plotting the pseudo-label positions without the transformation, \ie, the difference  between \eqnref{equ:flip} and \eqnref{equ:newflip}. 
Both approaches share the same initial pseudo-labels. For reference, we also compute the ground truth labels using the calibrated screen pose for flipped images.

As shown in  \figref{fig:trajectory}, the initial pseudo-labels have a significant offset from the ground truth. However, our method dynamically updates the pseudo-labels based on the fine-tuned network, progressively aligning them closer to the ground truth with each iteration. By the end of training, the pseudo-labels have only minimal offsets from the ground truth, demonstrating the effectiveness of our approach.
In contrast, the strategy without transformation fails to produce reliable pseudo-labels, leading to consistently large offsets from the ground truth. 

\section{Conclusion and Discussion}
In this work, we introduce a novel cross-task few-shot 2D gaze estimation method. By leveraging few-shot 2D samples, we adapt a 3D gaze model to 2D gaze estimation on unseen devices. Since the 3D gaze network is trained in 3D space without being tied to specific devices, it theoretically maintains robust performance across different platforms. Our experiments validate this by proving results on three datasets. Besides, the adaption is rapid and source-free, significantly broadening its practical applicability.

\noindent \textbf{Limitation:} Our method infers 2D gaze through mathematical derivation within the differentiable projection module. While this approach enhances model interpretability and reliability, it can occasionally result in failure cases. For instance, when the input images lack visible faces, the predicted 3D gaze can become erratic. In such scenarios, the intersection point between the 3D gaze vector and the screen plane may significantly deviate from the ground truth. This issue arises because, unlike neural networks that constrain outputs to a plausible range, a purely mathematical projection may yield extreme values, \eg, when the 3D gaze is nearly parallel to the plane. Although these cases can be easily flagged in real-world applications, they may introduce biases during evaluation.

\noindent \textbf{Future Directions:} 
In this paper, we address the challenge of 2D pseudo-labeling. However, several open questions remain. For instance, can we leverage unlabeled face images to further enhance performance? Traditional methods often utilize a standard calibration pattern, could we incorporate a similar strategy? It is worth noting that our approach requires collecting samples initially, akin to a calibration process. We argue that this step is essential as it provides the necessary anchors for adapting to unseen devices. Nonetheless, exploring user-unaware calibration techniques is also a promising direction for future research.

\newpage
\section{Acknowledgments}

This work was supported by the Institute for Information \& communications Technology Promotion(IITP) grant funded by the Korea government(MSIP) (No.RS-2024-00397615, Development of an automotive software platform for Software-Defined-Vehicle (SDV) integrated with an AI framework required for intelligent vehicles), as well as the Ramsay Research Fund, UK.

{
    \small
    \bibliographystyle{ieeenat_fullname}
    \bibliography{main}
}
%\input{sec/X_suppl}
% WARNING: do not forget to delete the supplementary pages from your submission 
% \input{sec/X_suppl}

\end{document}